\documentclass[conference]{IEEEtran}
\IEEEoverridecommandlockouts
\usepackage{cite}
\usepackage{amsmath,amssymb,amsfonts}
\usepackage{algorithmic}
\usepackage{graphicx}
\usepackage{textcomp}
\usepackage{comment}
\usepackage{xcolor}
\usepackage{url}

\def\BibTeX{{\rm B\kern-.05em{\sc i\kern-.025em b}\kern-.08em
    T\kern-.1667em\lower.7ex\hbox{E}\kern-.125emX}}
\begin{document}

\title{Prototype-Based Learning for Healthcare: \\A Demonstration of Interpretable AI}

%% Option 1

% \author{
% \IEEEauthorblockN{
% Ashish Rana,
% Ammar Shaker,
% Sascha Saralajew}
% \IEEEauthorblockA{
% \textit{NEC Laboratories Europe GmbH } \\
% \textit{Heidelberg, Germany} \\
% \{Ashish.Rana, Ammar.Shaker, Sascha.Saralajew\}@neclab.eu}
% \and
% \IEEEauthorblockN{
% Takashi Suzuki,
% Kosuke Yasuda,
% Shintaro Kato}
% \IEEEauthorblockA{
% \textit{NEC Solution Innovators, Ltd.} \\
% \textit{Tokyo, Japan} \\
% \{suzuki.tks,yasuda.kosuke,katou-s-mxn\}@nec.com}
% \and
% \IEEEauthorblockN{
% Toshikazu Wada, 
% Toru Kikutsuji}
% \IEEEauthorblockA{
% \textit{Kurashiki Central Hospital Preventive} \\
% \textit{Healthcare Plaza, Okayama, Japan} \\
% \{tw15651, tk11420\}@kchnet.or.jp}
% \and
% \IEEEauthorblockN{
% Toshiyuki Fujikawa}
% \IEEEauthorblockA{
% \textit{Kurashiki Central Hospital} \\
% \textit{Okayama, Japan} \\
% toshi@kchnet.or.jp}
% }

%% Option 2

% \author{
% \IEEEauthorblockN{
% Ashish Rana$^1$,
% Ammar Shaker$^1$,
% Sascha Saralajew$^1$,
% Takashi Suzuki$^2$,
% Kosuke Yasuda$^2$,
% Shintaro Kato$^2$, \\
% Toshikazu Wada$^3$,
% Toru Kikutsuji$^3$,
% Toshiyuki Fujikawa$^4$}

% \IEEEauthorblockA{
% $^1$NEC Laboratories Europe GmbH, Heidelberg, Germany\\
% $^2$NEC Solution Innovators, Ltd., Tokyo, Japan\\
% $^3$Kurashiki Central Hospital Preventive Healthcare Plaza, Okayama, Japan\\
% $^4$Kurashiki Central Hospital, Okayama, Japan\\
% \{Ashish.Rana, Ammar.Shaker, Sascha.Saralajew\}@neclab.eu,
% \{suzuki.tks, yasuda.kosuke, katou-s-mxn\}@nec.com,\\
% \{tw15651, tk11420\}@kchnet.or.jp,
% toshi@kchnet.or.jp}
% }

%% Option 3

\author{
\IEEEauthorblockN{
Ashish Rana$^1$, Ammar Shaker$^1$, Sascha Saralajew$^1$, Takashi Suzuki$^2$, Kosuke Yasuda$^2$, Shintaro Kato$^2$,\\
Toshikazu Wada$^3$, Toshiyuki Fujikawa$^4$, Toru Kikutsuji$^3$}
\IEEEauthorblockA{
$^1$NEC Laboratories Europe GmbH, Heidelberg, Germany;
$^2$NEC Solution Innovators, Ltd., Tokyo, Japan;\\
$^3$Kurashiki Central Hospital Preventive Healthcare Plaza and $^4$Kurashiki Central Hospital, Okayama, Japan}
\IEEEauthorblockA{
\{Ashish.Rana, Ammar.Shaker, Sascha.Saralajew\}@neclab.eu,
\{suzuki.tks, yasuda.kosuke, katou-s-mxn\}@nec.com,\\
\{tw15651, tk11420\}@kchnet.or.jp, toshi@kchnet.or.jp}
}

\maketitle

\begin{abstract}
%\url{https://www3.cs.stonybrook.edu/~icdm2025/index.html} \\
%\url{https://icdm25demo.github.io/} \\
%The 13th Workshop on Data Mining in Biomedical Informatics and Healthcare (DMBIH’25)
%https://www.oakland.edu/secs/dmbih-workshop-2025/ \\
%https://www.strobe-statement.org/ \\
%\textcolor{blue}{Problem Introduction: Interpretability need for disease risks and recommendation assistance in Healthcare \\
%Prototypes Introduction: Unique advantages offered by prototypes based learning \\
%System Introduction: Functional advantages of the proposed system through prototypes and %available downstream features \\ 
%System Performance: SoTA performance compared to Cox model and alongside captures  biological disease trends \\}
Despite recent advances in machine learning and explainable AI, a gap remains in personalized preventive healthcare: predictions, interventions, and recommendations should be both understandable and verifiable for all stakeholders in the healthcare sector. We present a demonstration of how prototype-based learning can address these needs. Our proposed framework, ProtoPal, features both front- and back-end modes; it achieves superior quantitative performance while also providing an intuitive presentation of interventions and their simulated outcomes.
\end{abstract}

% ### Abstract
% Problem Introduction: Interpretability need for disease risks and recommendation assistance in Healthcare
% Prototypes Introduction: Unique advantages offered by prototypes based learning
% System Introduction: Functional advantages of the proposed system through prototypes and available downstream features
% System Performance: SoTA performance compared to Cox model and alongside captures biological disease trends


\section{Introduction}

\begin{comment}

Here, we start talking about XAI in general and then introduce prototypes.
More specifically, we introduce prototypes. After that, we introduce health checkup centers:
\begin{enumerate}
\item XAI/interpretability in general, specific discussion on prototype-based learning
\item Health checkup centers using AI systems, existing AI in healthcare approaches and possible limitations
\item Using prototype-based learning to capture biological trends with performance improvements, improving interventions for proposed lifestyle changes, and wrapping all the functionalities in the form of a meaningful demo.
\end{enumerate}
\end{comment}

The great performance of deep learning methods comes with the challenge of mistrust in their decisions, caused by the lack of human understanding of how results are derived. While Explainable AI (XAI) has emerged to address these challenges, most post-hoc approaches such as LIME~\cite{ribeiro2016should} and SHAP~\cite{lundberg2017unified} focus on bridging the gap between black-box models and human reasoning, without making the decision process itself inherently interpretable~\cite{rudin2019stop}.
Prototype-Based Learning (PBL) methods, on the other hand, are naturally interpretable: they learn prototypes that originate from the input space or can be easily projected onto it, and rely on simple classification rules that embed human-comprehensible distance measures~\cite{biehl2016prototype}.

Health checkup centers in Japan offer the public annual checkups. These centers serve the purpose of disease prevention for the majority of healthy people, early disease identification, or disease control in advanced stages. The results are often summarized in tabular form with calculations of different disease risks, which are difficult for non-experts to interpret. In such cases, medical operators must devote time to explaining the results to patients and providing preventive or mitigating recommendations. This process is not only cumbersome but also consumes the valuable time of doctors and medical staff. Having AI systems that are both predictive and inherently explainable becomes vital to overcoming these challenges. While prototype-based learning is a good candidate to address this problem, it must be carefully adapted and treated to suit the needs of non-AI operators.

Carloni et al.~\cite{carloni2022applicability} show that Prototypical Part Networks (ProtoPNet)~\cite{chen2019looks} are capable of learning and classifying breast tumors.
Xu et al.~\cite{xu2024protomix} introduce ProtoEHR, a hierarchical prototype learning framework for electronic healthcare records (EHRs), designed for prediction tasks such as mortality and readmission.
While these works demonstrate the effectiveness of PBL as interpretable predictors, their potential to serve as digital twins remains unexplored, particularly in relation to providing interventive recommendations for disease prevention.

Recent advances in AI simulation have enabled the visualization of near-future projections of numerical values and health risks (\cite{fujihara2023machine}, \cite{yasuda2024comprehensive}), offering individuals a clearer understanding of their potential health trajectories. Motivated by this, we explore the potential of applying a prototype-based learning framework to support such simulations in a personalized and interpretable manner. In particular, we introduce a novel representation grounded in the concept of an “ideal healthy self.” This approach would be innovative in that it shifts the reference point from abstract statistical norms to a personalized health ideal. We hypothesize that this personalization enhances the sense of ownership for the examinee, thereby increasing motivation to engage with preventive actions and lifestyle improvements. To investigate the feasibility of this approach, we constructed a demonstration environment that enables interactive simulation and interpretation based on PBL.

This work introduces ProtoPal\footnote{ProtoPal is available to reviewers at \url{https://ienlaight.jedis.nlehd.de}
 (User: \texttt{ienlaight\_user}, Pass: \texttt{z2Ud6\_+73w-kk9H=})
 }, an interpretable and interactive framework for training on diseased and healthy individuals. It is able to predict and capture the biology of diseases, as well as generate interventions and optimal recommendations via the simulation of individuals' digital twins.
\begin{figure*}
 \center
  \includegraphics[width=.825\textwidth]{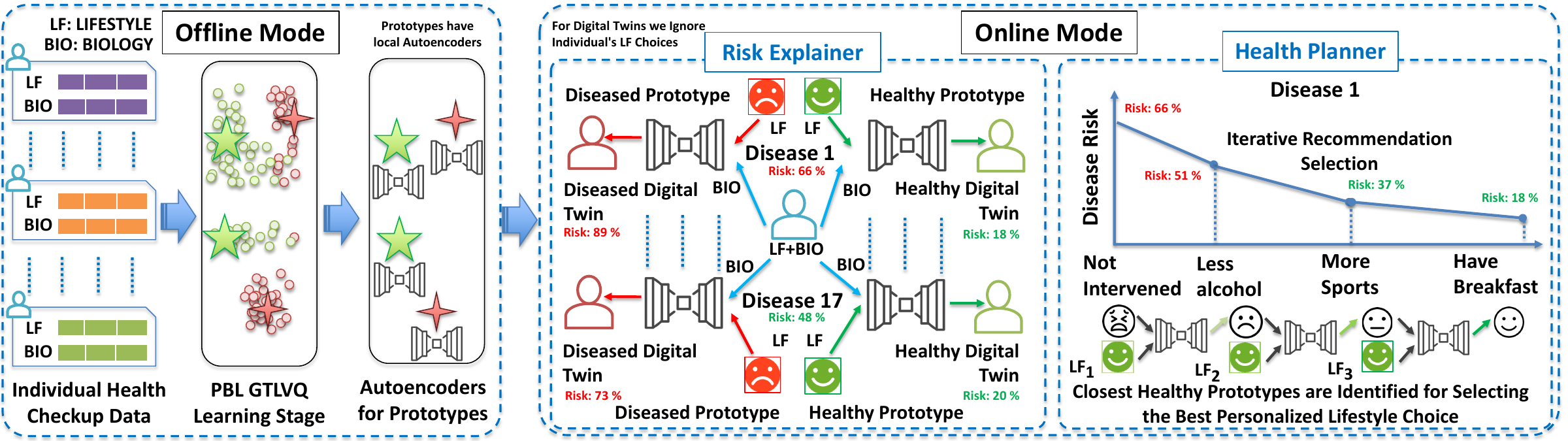}
  \caption{The architecture of ProtoPal. \textit{Left:} The offline mode, which includes training GTLVQ and fitting denoising autoencoders. \textit{Right:} The online mode, which first features a risk explainer that predicts an individual’s risk score. And second, a health planner that recommends a trajectory of lifestyle changes and illustrates how simulated biological vitals evolve through a series of intermediate healthier digital twins.}
  \label{fig:figure_platform_pipeline}
\end{figure*}

\section{Kurashiki Central Hospital}
\begin{comment}
In this section, we introduce the KCH center and its objective. Thereafter, we explain the data, visits and diseases.
\begin{enumerate}
\item KCH Introduction
\item KCH data: number of patients, ICD, visits
\item attributes and lifestyle features
\item the target of the data: recommendations and interventions
\end{enumerate}
\end{comment}

This work is inspired by and built on the health checkup data from Kurashiki Central Hospital (KCH). It has been reviewed and approved by the KCH review board (Approval No. 4130). Further, our data analysis procedures strictly follow and comply with the GDPR regulations for collecting and processing personal data, as extensively elaborated in Data Compliance with GDPR\footnote{Regulation (EU) 2016/679 of the European Parliament and of the Council of 27 April 2016 on the protection of natural persons with regard to the processing of personal data and on the free movement of such data, and repealing Directive 95/46/EC (General Data Protection Regulation) (Text with EEA relevance).}. The dataset includes health checkup records from 92,174 individuals who visited KCH between January 2012 and September 2022. Although many participants have multiple records corresponding to repeated checkup visits, in this study we consider only the first visit for each individual. At the first health checkup, collected factors included demographics (age, sex, height, weight, BMI, waist circumference); labs outcomes (blood pressure, liver enzymes, HDL, high density lipoprotein cholesterol (HDL), low density lipoprotein cholesterol (LDL), triglycerides, glucose, HbA1c, uric acid); lifestyle (smoking, activity, diet, alcohol, sleep); medical history (stroke, heart, kidney disease, anemia); and medication use (hypertension (HT), diabetes mellitus (DM), and hyperlipidemia (HL)). Lifestyle, medical history, and medication were assessed by questionnaire; see~\cite{yasuda2024comprehensive} for details. Diseases are encoded using ICD-10 medical classification; in this study, we focus only on 17 frequent diseases listed in the first column of Table~\ref{tbl:model_comparison_results}.

\section{Framework Structure}

%The platform has offline mode for data pre-processing, model training and result validation. Figure \ref{figure_platform_pipeline} represents the offline mode platform pipeline architecture for the \textit{i}enAIght. Further, the online mode provides an interactive Streamlit frontend to get active personalized healthcare information. Additionally, in this online mode the backend executes the model for inference and conducts recommendation computations.

\begin{figure}
 \center  
  \includegraphics[width=0.45\textwidth,trim={0 0 0 43}, clip]{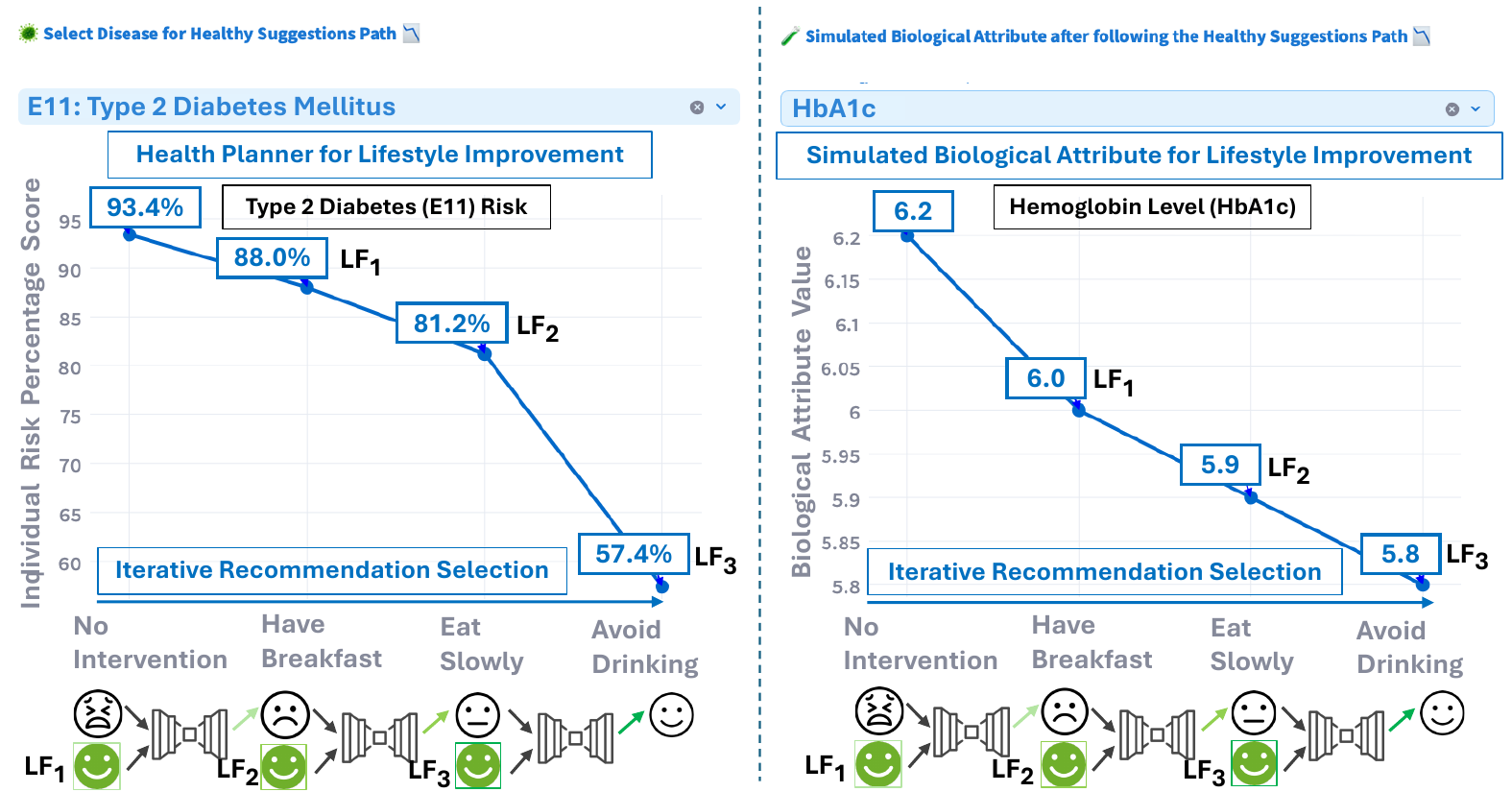}
  \caption{
  % The health planner provides personalized iterative healthy lifestyle recommendations. It does that by using the closest healthy prototype to identify the best possible lifestyle recommendation for best improvement at each step.
  The health planner for the same highly diseased old individual provides personalized iterative healthy lifestyle recommendations.
  This is done by using the closest healthy prototype to identify the best possible lifestyle recommendation for best disease risk reduction at each step.
  Also, first improvements in relatively easier to follow healthy eating habits are recommended.
  And reduction in harder or addictive activities like alcohol consumption is recommended in later stages for the elderly individual.
  }
  \label{fig:Health_Planner}
\end{figure}

Our framework, ProtoPal, consists of two major modes. The offline mode learns prototypes for disease classification and fits autoencoders in their vicinity to enable simulation functionalities.
The online mode operates on query individuals\footnote{All results depicted in Figures~\ref{fig:figure_platform_pipeline},\ref{fig:risk_explainer} and~\ref{fig:Health_Planner} are based on synthetically generated individual data.}: their risk affinity for each studied disease is computed and explained in the Risk Explainer using both the closest healthy and diseased prototypes. Moreover, an interactive Streamlit frontend provides personalized lifestyle recommendations along with their estimated effects on overall risk. The framework facilitates the simulation of changes in an individual’s vitals when adopting some or all lifestyle habits of a prototype. The resulting simulated individual is referred to as a healthy or diseased digital twin, depending on whether the lifestyle of a healthy or diseased prototype is adopted.
Figure~\ref{fig:figure_platform_pipeline} illustrates the architecture of the framework.

\subsection{Offline Mode: Interpretable Prototype-based Learning}
The offline mode triggers the learning of prototypes from the enlAIght library, a proprietary technology of NEC Labs Europe\footnote{The repository of enlAIght is \url{https://github.com/nec-research/enlaight}}. Prototypes, often, are examples from the input space learned in a distance-based fashion for classification problems.  
Given a dataset $D = \{(x, c) \mid x \in \mathbb{R}^d, c \in \mathcal{Y}\}$, with class label $c \in \{D: \text{diseased}, H: \text{healthy}\}$, Generalized Learning Vector Quantization (GLVQ) learns a set of prototypes  
\[
W = \{(w_j, c_j) \mid w_j \in \mathbb{R}^d, \; c_j \in \mathcal{Y}, \; j = 1, \dots, N \},
\]  
where each prototype $w_j$ is associated with a label $c_j$.  
The learning objective minimizes a cost function based on the relative distance of an individual $x$ to the closest prototype of the same class versus the closest prototype of a different class.  
Different GLVQ variants employ different distance measures, such as Euclidean, Mahalanobis (GMLVQ) ~\cite{sato1995generalized}, or tangent distance (GTLVQ) ~\cite{saralajew2016adaptive}.  
In this demo, we assume the GTLVQ variant.

Prototypes usually have a simple classification rule; however, aside from the crisp prediction score, risk estimators are often evaluated based on their correct ranking of participants, either using the Area Under the Curve (AUC) or the C-index in the presence of censoring.  
To equip prototypes with the ability to compute risk scores, we propose assigning a continuous risk score to each participant rather than a discrete class label.  
For an individual $x$ and the set of prototypes $(P,c)$ with class $c \in \{D,H\}$, we find the smallest hypersphere $M(x)$ (according to the distance measure $m$) around $x$ that includes prototypes from both classes. Let $B = \{(P,c) \in W \mid P \in M(x)\}$ denote the prototypes inside this hypersphere. The risk of being a diseased sample is computed as follows:  
\begin{align}
r_0(x) = \frac{\sum_{(P_i, c_i) \in B \wedge c_i = 0} 1/m(x,P_i)}{\sum_{(P_i, c_i) \in B} 1/m(x,P_i)} \ .
\end{align}
In this way, the density and class homogeneity of prototypes in the neighborhood determine the belief that the individual belongs to the diseased or healthy class. Hence, the neighborhood having more diseased prototypes will have a higher risk score for the individual and vice versa. 

Finally, as the last step of offline learning, we fit a denoising autoencoder $E_c$ on participants around each prototype $(P,c)$ for the purpose of intervention simulation, explained later.

\begin{figure}
 \center
  \includegraphics[width=0.475\textwidth, trim={0 12 0 10}, clip]{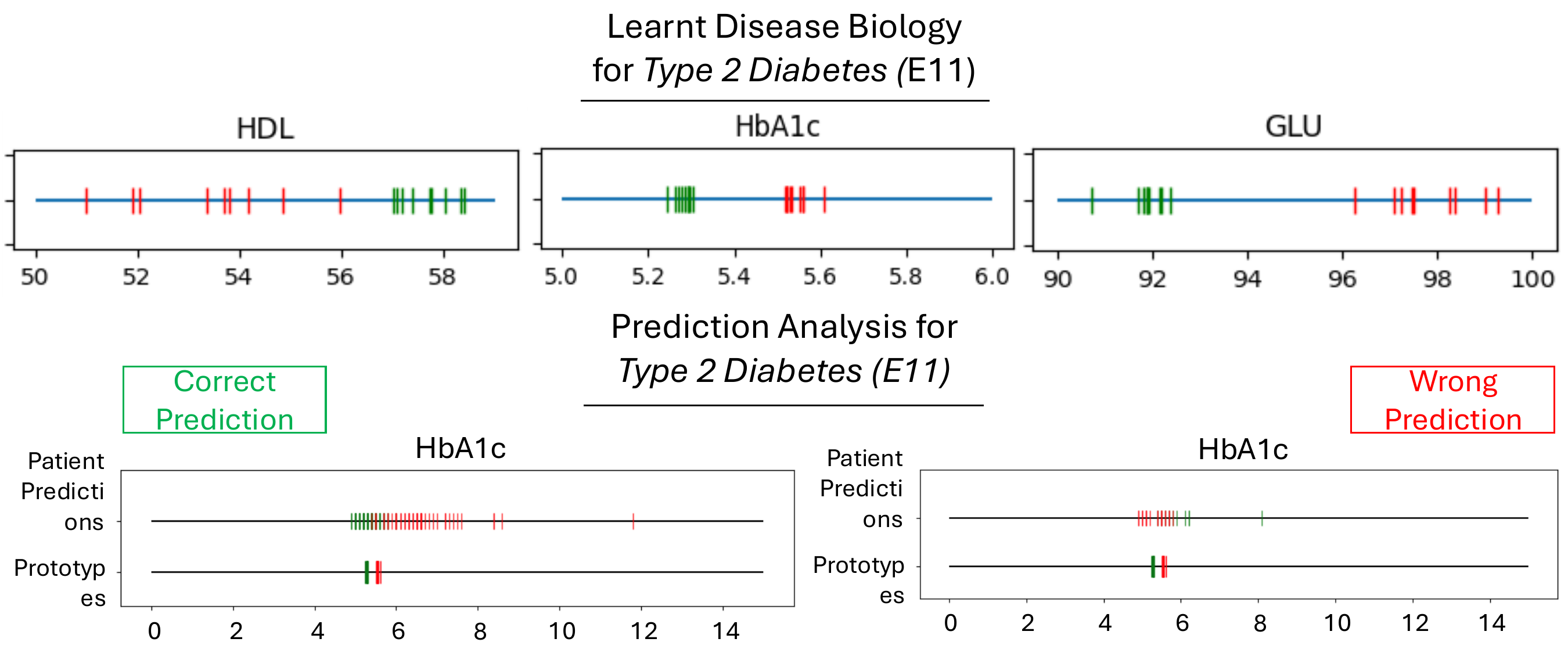}
  \caption{The continuous dot plots with line legends demonstrate that the learnt models capture disease biology \textit{(top row)} and assists in identifying reasoning behind correct or incorrect predictions \textit{(bottom row)}.}
  \label{fig:biological_disease_trends_and_prediction_analysis}
\end{figure}

\textbf{Quantitative Evaluation:} Before employing the learned prototypes for lifestyle interventions and recommendations, we must first verify that they capture sufficient disease characteristics to distinguish diseased from healthy individuals across the studied diseases. This problem may be affected by censoring, since individuals who appear healthy may still be at high but undiscovered risk. Therefore, we evaluate the ranking of the predicted risk scores and compare it with that of the Cox proportional hazards model~\cite{Cox:1972:RMLT}.
Table~\ref{fig:biological_disease_trends_and_prediction_analysis} shows that PBL outperforms the Cox model in 12 out of 17 diseases in terms of higher AUC.

\textbf{Qualitative Assessments:} 
%Beyond superior performance, the learned prototypes also provide a rich source of information by capturing biological disease trends. 
The offline mode also provides assistance in the qualitative assessment of disease risk prediction models. The first-row subplots in Figure~\ref{fig:biological_disease_trends_and_prediction_analysis} illustrate how prototypes distribute across different health vitals for Type 2 Diabetes Mellitus (E11). Diseased prototypes \textit{(red lines)} exhibit higher values for risk-related attributes such as bad cholesterol (HDL), hemoglobin levels (HbA1c), and glucose (GLU), whereas healthy prototypes \textit{(green lines)} show the opposite trend. The second-row subplots in Figure~\ref{fig:biological_disease_trends_and_prediction_analysis} display correct and incorrect predictions, demonstrating how prototypes can also help explain the sources of misclassification.
For example, incorrectly predicted diseased individuals \textit{(red lines)} have HbA1c values similar to those of healthy prototypes \textit{(green lines)}, and conversely, misclassified healthy individuals \textit{(green lines)} resemble diseased prototypes.

%\textbf{Qualitative Assessments:} 
%The developed tool with risk explainer and health planner modules provides an interactive interface for qualitative assessment of the developed disease risk prediction models.
%Firstly, Figure \ref{fig:biological_disease_trends_and_prediction_analysis} highlights that the prototypes learned by the models capture disease biology in the first-row subplots.
%Where unhealthier prototypes \textit{(red lines)} have higher value for bad cholesterol (HDL), hemoglobin levels (HbA1c), glutamate (GLU) attributes, and vice versa for healthy prototypes \textit{(green lines)}.
%Furthermore, the second-row subplots of correct and incorrect predictions in Figure \ref{fig:biological_disease_trends_and_prediction_analysis} demonstrate that through prototypes we can also decode the reason behind our incorrect prediction.
%For example, for incorrectly predicted unhealthy samples \textit{(red lines)} have a Hb1Ac value similar to healthier prototypes \textit{(green lines)} and vice versa for incorrectly predicted healthy samples \textit{(green lines)}.

\subsection{Online Mode: Interactive Personalized Healthcare Tool}

The online mode contains two main frontend functionalities, facilitated by the previously learned prototypes and the fitted autoencoders.  
In this mode, the medical operator enters the vitals of an individual (age, gender, body mass index (BMI), ...) and lifestyle features (smoking, drinking, ...) and then computes and explain the risk using the risk explainer, or additionally provides a health improvement plan.

\begin{figure*}
 \center
  \includegraphics[width=0.85\textwidth]{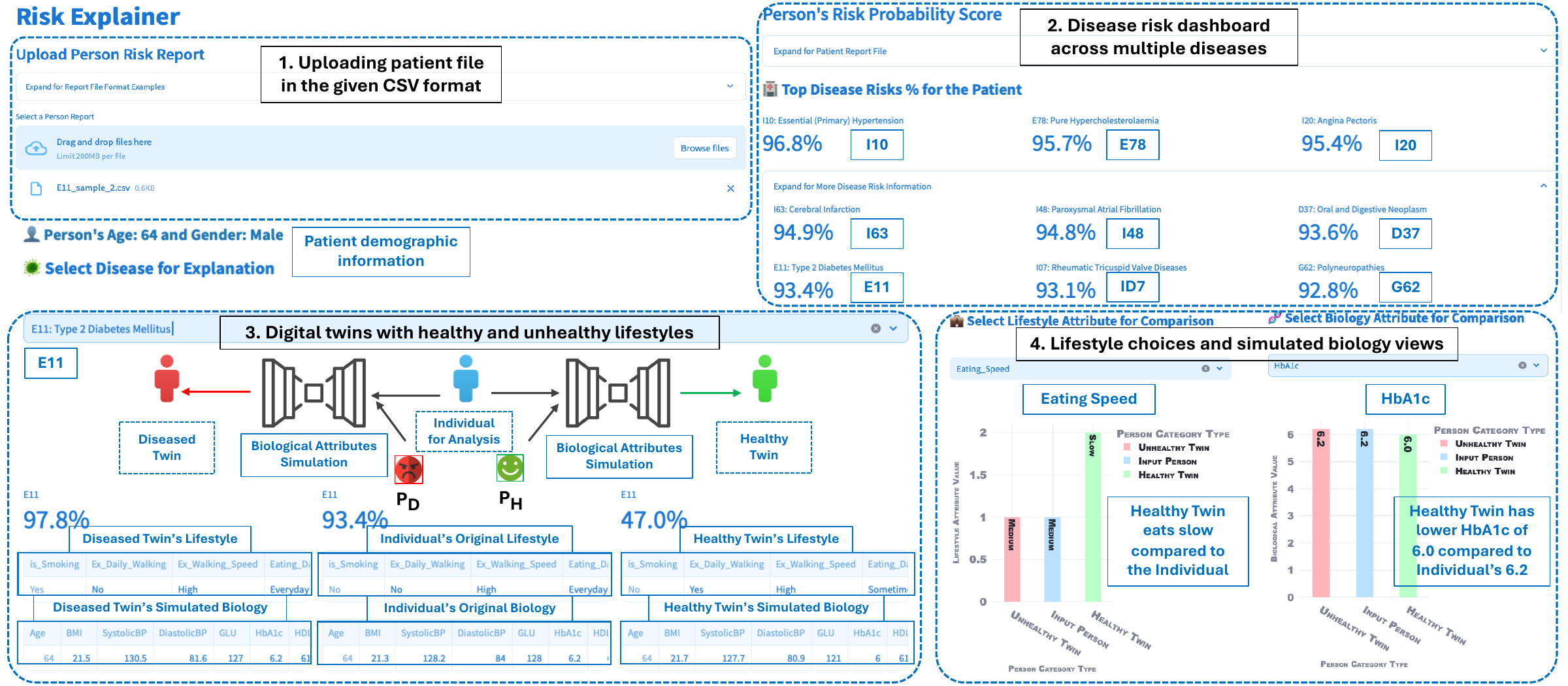}
  \caption{
  % The risk explainer output for a given patient with Type-2 Diabetes (E11) disease risk, where the important disease risk determining features have unsafe values for the unhealthy digital twin and vice-versa. Further, healthier lifestyle choices which the healthier digital twin is following are also highlighted with the corresponding disease risk decrease and vice-versa.
  The risk explainer output for a given highly diseased old individual, where disease risk dashboard summarizes all the different disease risks amongst various diseases.
  Additionally, detailed analysis of Type-2 Diabetes Mellitus (E11) is provided as well with the assistance from healthy and diseased digital twins.
  These digital twins are respectively created from the closest healthy \textit{($P_H$)} and closest diseased prototypes \textit{($P_D$)} after wholly adopting the respective lifestyle and correspondingly simulating all the biological attributes.
  Finally, the detailed lifestyle choices and simulated biology comparison bar plots contrast the individual's attributes against diseased and healthy digital twins.
}
  \label{fig:risk_explainer}
\end{figure*}

\textbf{Risk Explainer:} Given the individual $x$, the medical operator selects one of the diseases available in the framework. The system then computes the nearest diseased prototype $P_D$ and the nearest healthy prototype $P_H$, which represent the person's closest diseased and healthy prototypes.  
As illustrated in Figure~\ref{fig:risk_explainer}, the risk of Type 2 Diabetes Mellitus (E11) for the participant $x$ is computed. In this example, $x$ shows a high tendency to develop diabetes (score of 91.4 \%). However, by adopting the lifestyle of the healthy prototype $P_H$, the risk would drop to approximately 47.0 \%, as confirmed by improvements in the simulated HbA1c hemoglobin test value generated by $P_H$'s autoencoder. Conversely, following the unhealthy habits of the diseased prototype $P_D$ leads to deteriorated biological vitals and an increased risk of diabetes.

\begin{comment}
\begin{enumerate}
\item Elaborating GTLVQ
\item Corresponding risk score computation, and modified AUC calculation formula
\item Simulation via autoencoders for the non-intervened features, and distance weighting hyperparameter for tuning risk scores given the closest healthy and unhealthy prototypes
\end{enumerate}

\begin{enumerate}
\item Online Mode: Risk Explainer and Health Planner % and Biology Explainer
\begin{enumerate}
\item Risk Explainer: This feature explains the risk of any disease for a given patient with the help of reference closest healthy and unhealthy prototypes.
\begin{enumerate}
\item The risk is across all the diseases listed in the results Table \ref{model_comparison_results} with higher risk diseases being presented first. And there are multiple disease specific models that calculate the probability of disease risk based on the above distance based risk calculation metric. 
\item Further, it provides the reference closest healthy and unhealthy prototypes which are made demographically consistent through autoencoder simulation by fixing the age and gender attributes. Also, detailed lifestyle attributes and simulated biological attributes for the closest unhealthy and healthy digital twin are provided for comparison and review as well.
\end{enumerate}
The above information assists in explaining the effect of healthy and unhealthy choices on different disease risk values and biological attributes through the assistance of a person's digital twins.
\end{enumerate}
\end{comment}

\textbf{Health Planner:} While the risk explainer allows for simulating lifestyle interventions by fully mimicking the behavior of the healthy prototype via copying all its lifestyle to generate the healthy digital twin, this approach is crude and unrealistic.  
Based on interviews with medical doctors, psychologists, and patients, we realized that people are generally reluctant to completely change their habits. Even sick individuals often resist drastic modifications despite deteriorating health. However, both healthy and diseased groups tend to be more cooperative when changes are limited, gradual, and well-planned.  
\textit{Limited} means that a single change can be acceptable, whereas many simultaneous changes may be overwhelming.  
\textit{Gradual} refers to progressive adjustments, such as reducing alcohol consumption from five times a week to twice a week, rather than enforcing complete abstinence immediately. The same principle applies to practicing sports, but in the opposite direction, where activity levels can be increased step by step.  

Planning such gradual lifestyle changes is the core purpose of the health planner.  
After identifying the corresponding healthy prototype for the individual $x$, the health planner evaluates each of its lifestyle factors and their effect on $x$, assuming $x$ agrees to adopt them. We employ a greedy algorithm that, at each step, selects the lifestyle change leading to the largest risk reduction.  
The resulting ordered list of changes forms a plan intended to protect healthy individuals from the disease and improve the condition of diseased individuals. Moreover, each step in the plan is accompanied by simulated impacts on the biological vitals of $x$, represented as intermediate healthy digital twins generated by the fitted autoencoders.
\begin{comment}
\begin{enumerate}
\item Health Planner: This provides iterative improvements in the lifestyle to lower a patient's disease risk.
\begin{enumerate}
\item This feature iteratively in a greedy manner finds the most effective recommendation and selects to calculate the new reduced risk.
And, this step is iteratively until all the lifestyle improvement recommendations are taken for the patient's health improvement.
\item Further, at each step when the lifestyle choice is selected we simulate the biological vitals as well for this new recommended lifestyle intervention. And, we continue to carry out this simulation at each step for accurate disease risk prediction and biological attributes variation analysis post the lifestyle recommendations.
\end{enumerate}
The above information assists in defining the most optimized recommendation plan to improve their health. And potentially this health planner can reduce the workload on primary healthcare providers with preemptive disease risk reduction.

\end{enumerate}

\end{comment}

% \section{Results}
\begin{comment}
\section{Quantitative Evaluation} 
\begin{enumerate}
\item Experiment setup and model training details, both GTLVQ and Autoencoder models 
\item GTLVQ's AUC comparison with Cox model
\end{enumerate}
\end{comment}

\begin{table}[ht]
\centering
\begin{tabular}{|c|l|c|c|}
\hline
\textbf{ICD10} & \textbf{Disease Names} & \textbf{Cox} & \textbf{GTLVQ} \\
\hline
I10 & Essential (primary) hypertension & 0.795 & \textbf{0.801} \\
I20 & Angina pectoris & 0.692 & \textbf{0.771} \\
I21 & Acute myocardial infarction & \textbf{0.779} & 0.774 \\
I42 & Cardiomyopathy & \textbf{0.880} & 0.800 \\
I48 & Atrial fibrillation and flutter & 0.813 & \textbf{0.840} \\
I50 & Heart failure & 0.738 & \textbf{0.786} \\
I60 & Subarachnoid haemorrhage & \textbf{0.830} & 0.710 \\
I61 & Intracerebral haemorrhage & 0.728 & \textbf{0.776} \\
I63 & Cerebral infarction & 0.779 & \textbf{0.806} \\
I70 & Atherosclerosis & 0.784 & \textbf{0.789} \\
E11 & Type 2 diabetes mellitus & 0.817 & \textbf{0.855} \\
E78 & Disorders of lipoprotein metabolism & 0.752 & \textbf{0.781} \\
K70 & Alcoholic liver disease & \textbf{0.960} & 0.936 \\
K74 & Fibrosis and cirrhosis of liver & 0.753 & \textbf{0.825} \\
K75 & Other inflammatory liver diseases & 0.629 & \textbf{0.709} \\
K76 & Other diseases of liver & 0.652 & \textbf{0.693} \\
N18 & Chronic kidney disease & \textbf{0.807} & 0.751 \\
\hline
Wins & &5&12\\
\hline
\end{tabular}
\caption{Comparison of Cox model vs. GTLVQ across 17 ICD10 diseases.}
\label{tbl:model_comparison_results}
\end{table}

\begin{comment}
\begin{figure*}
 \center  \includegraphics[width=0.975\textwidth]{health_explainer_module.pdf}
  \caption{The health planner provides personalized iterative healthy lifestyle recommendations. It does that by using the closest healthy prototype to identify the best possible lifestyle recommendation for best improvement at each step.}
  \label{tbl:risk_explainer}
\end{figure*}
\end{comment}

%\section{Future Work}
% \begin{enumerate}
% \item Addressing some limitations of existing work, like learnt healthy or disease prototypes are all at decision boundary
% \item We intend on extending our analysis to time-based interventions as well
% \end{enumerate}
% \section{Implementation details: Demos must be based on fully implemented and tested systems.} 

%The most important extension of this work will be the addition of temporal interventions to predict the risk of onset of the disease. As observed from models trained in the offline mode, healthy and unhealthy prototypes cluster close to the decision boundary across all features. Future work will explore constraint-based approaches to enhance prototype diversity, enabling the capture of distinct patient cohorts with shared clinical characteristics. Finally, as part of our future qualitative assessment effort through experts, we would focus on getting this system tested with future disease risk predictions. 

\section{Conclusion and Outlook}

In this demonstration, we present an interactive framework for applying prototype-based learning in the healthcare sector. To our knowledge, this is the first work to demonstrate how prototypes can be used to recommend interventions, estimate risk reduction trajectories when these interventions are followed, and simulate improvements in health vitals. The proposed framework also sets the groundwork for future case studies exploring different prototype learning methods. Currently, the framework is being extended to allow medical personnel to provide feedback and corrections on the learned prototypes, as well as to incorporate patient feedback on the quality of interventions and health plans into the offline learning phase. In addition, our goal is to explore constraints-based approaches to enhance prototype diversity and better capture distinct patient cohorts with shared clinical characteristics.

\bibliographystyle{IEEEtran}
\bibliography{library}

\end{document}